\documentclass{article}

\PassOptionsToPackage{numbers, compress}{natbib}
\usepackage[final]{neurips_2019}

\usepackage[utf8]{inputenc} % allow utf-8 input
\usepackage[T1]{fontenc}    % use 8-bit T1 fonts
\usepackage{hyperref}       % hyperlinks
\usepackage{url}            % simple URL typesetting
\usepackage{booktabs}       % professional-quality tables
\usepackage{amsfonts}       % blackboard math symbols
\usepackage{nicefrac}       % compact symbols for 1/2, etc.
\usepackage{microtype}      % microtypography

\usepackage[ruled, vlined,linesnumbered]{algorithm2e}
\usepackage[dvipsnames]{xcolor}
\usepackage{bm}
\usepackage{amsmath}
\usepackage{amssymb}
\usepackage{tikz}
\usetikzlibrary{shapes.geometric, arrows, shadows, bayesnet}
\usepackage{tabularx}
\usepackage{amsthm}

\newtheorem{definition}{Definition}

\DeclareMathOperator*{\argmax}{arg\,max}

\makeatletter
\newcommand*{\indep}{\mathbin{\mathpalette{\@indep}{}}}
\newcommand*{\nindep}{\mathbin{\mathpalette{\@indep}{\not}}}
\newcommand*{\@indep}[2]{\sbox0{$#1\perp\m@th$}\sbox2{$#1=$}\sbox4{$#1\vcenter{}$}
\rlap{\copy0}\dimen@=\dimexpr\ht2-\ht4-.2pt\relax\kern\dimen@{#2}\kern\dimen@\copy0} 
\makeatother

\newcommand{\ra}[1]{\renewcommand{\arraystretch}{#1}}
\usetikzlibrary{arrows,positioning} 
\tikzset{
    >=stealth',
    punkt/.style={
           rectangle,
           rounded corners,
           draw=black, very thick,
           text width=6.5em,
           minimum height=2em,
           text centered},
    pil/.style={
           ->,
           thick,
           shorten <=2pt,
           shorten >=2pt,}
}

\title{Optimal experimental design via Bayesian optimization: active causal structure learning for Gaussian process networks}

\author{%
  Julius von K\"ugelgen \\
  MPI for Intelligent Systems\\
  University of Cambridge \\
  \texttt{jvk@tuebingen.mpg.de}
 \And 
 Paul Rubenstein\\
 MPI for Intelligent Systems\\
 University of Cambridge \\
  \texttt{paul.rubenstein@tuebingen.mpg.de}
  \And
  Bernhard Sch\"olkopf\\
  MPI for Intelligent Systems\\
  Amazon\\
  \texttt{bs@tuebingen.mpg.de}
  \And
  Adrian Weller\\
  University of Cambridge\\
  The Alan Turing Institute\\
  \texttt{aw665@cam.ac.uk}
 }

\begin{document}

\maketitle

\begin{abstract}
We study the problem of causal discovery through targeted interventions.
Starting from few observational measurements, we follow a Bayesian active learning approach to perform those experiments which, in expectation with respect to the current model, are maximally informative about the underlying causal structure.
Unlike previous work, we consider the setting of continuous random variables with non-linear functional relationships, modelled with Gaussian process priors.
To address the arising problem of choosing from an uncountable set of possible interventions, we propose to use Bayesian optimisation to efficiently maximise a Monte Carlo estimate of the expected information gain.
\end{abstract}

\section{Introduction and motivation}
Many of the broad ranging capabilities of human cognition---as opposed to the rather narrow intelligence of current AI systems---can be attributed to the possession of an internal world model \citep{lake2017building}.
These include (among others) the abilities to explain one's decisions, to transfer previously acquired knowledge, to efficiently adapt to new situations, and to imagine alternative futures.
Such internal models are thought to represent causal relationships between real-world entities and concepts.
There is evidence that such causal models are acquired already early in childhood, largely through playful interaction with the environment, and that they are continuously updated throughout life in light of new observations.
This approach to learning about the world through experimenting, observing evidence, and subsequently updating our hypotheses shows striking similarities to the scientific method---an idea that is termed the ``child as a scientist'' \citep{gopnik1996scientist}.

We propose to draw inspiration from the human approach to building world models by simulating an intelligent agent which repeatedly interacts with its environment.
Specifically, we consider an unsupervised setting where, in absence of a clear task or external reward, the agent's aim is to learn a causal model of its environment through targeted experimentation.
For example, we may think of a robot exploring its surroundings, a virtual agent in a simulated environment, or a scientist in a lab.

\section{Problem setting}
Formally, we consider an environment characterised by a structural causal model (SCM) \cite{Pearl2009} over a set of $d$ real-valued observable variables $\mathbf{X}=\{X_1,...,X_d\}$.
We assume that the corresponding causal graph $G^*$ is a directed acyclic graph (DAG), and that the functional relationships between each variable $X_i$ and its  causal parents $\mathbf{Pa}_i^{G^*}$ can be captured by an additive noise model (ANM) \cite{hoyer2009nonlinear},
\begin{equation}
    \label{eq:ANM}
    X_i = f_i(\mathbf{Pa}_i^{G^*})+\epsilon_i, \quad \quad (i=1,...,d)
\end{equation}
where $\epsilon_i$ are mutually independent noise terms (i.e., we assume acyclicity, causal sufficiency, and additive noise).
The induced observational distribution $P$ then factorises according to $G^*$.

\begin{figure}[]
    \centering
    \begin{tikzpicture}[node distance=1cm, auto,]
        \node[punkt] (agent) {Intrinsically Motivated Agent};
         %edge[pil,bend left=45] (environment)
        \node[punkt, right=2cm of agent] (environment) {Environment: Causal Model over $X_1,...,X_d$};
        % \path[->] (agent)  edge [bend left=45]  node[top]  {$a=1$}         (environment)        
        \path (agent.north) edge [pil,bend left=45] node[text width=12em,text centered]  {chooses an experiment $a_t := do(X_{j}=x)$} (environment.north);
        \path (environment.south) edge [pil,bend left=45] node[text width=12em,text centered]  {provides a sample\\ $D_t\sim P\big(\mathbf{X}_{-j}|do(X_j=x)\big)$}(agent.south);
    \end{tikzpicture}
    \caption[Schematic of active Bayesian causal discovery]{Illustration of the proposed setting for active Bayesian causal discovery. An intrinsically motivated agent aims to build a causal world model by repeated and targeted interaction with its environment. The latter is represented as a causal model over a set of observable variables, and can be queried by the agent by means of intervening on some of the variables.}
    \label{fig:ABCD}
\end{figure}
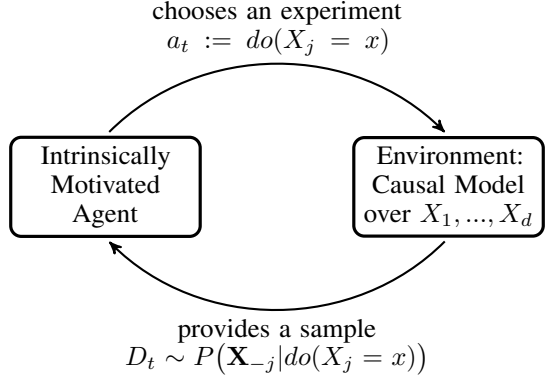

At each timestep $t=1,...$ the agent can perform an experiment $a_t$ which consists of intervening on one of the variables and fixing its value, $do(X_j=x)$.
Next, it observes the outcome of the experiment in form of a sample of the other variables from the corresponding interventional distribution, $P(\mathbf{X}_{-j}|do(X_j=x))$.
Finally, the agent uses the observed outcome to update its beliefs in order to plan what experiment best to perform next.
This process is repeated a finite number of times, or until the agent has reached sufficient confidence in its causal world model, see figure \ref{fig:ABCD}.

\section{Active Bayesian causal discovery}
As opposed to causal discovery methods from observational data \cite{mooij2016distinguishing, Spirtes1993}, our setting differs in that we aim to actively learn causal structure and functional relationships in a sequential (online) fashion using targeted interventions.
In order to make informed choices about which experiments to perform next it is valuable to keep track of uncertainties in current beliefs.
Maintaining uncertainty estimates over both graph structures and their associated parameters may also help to deal with an additional trade-off:
complex behaviour of a variable may be explained either by simple dependencies on many parent variables, or alternatively by complex dependence on only one or a few variables.

We thus adopt a Bayesian approach to causal discovery 
\citep{heckerman1995bayesian, heckerman2006bayesian}.
In the Bayesian framework, we start with a prior distribution $P(G)$ over possible causal graphs $G\in\mathcal{G}$ (where $\mathcal{G}$ is the set of all DAGs over $d$ variables).
For each graph $G$, we also place a prior $P(\theta_G|G)$ over its associated parameters $\theta_G\in \Theta_G$.
Our prior beliefs thus factorise as $P(G,\theta_G)=P(G)P(\theta_G|G)$.
Finally, we define a likelihood function $P(\mathbf{D}|\theta_G,G)$ which describes how to generate data $\mathbf{D}$ from the causal model encoded by the pair $(G,\theta_G)$.
The marginal likelihood, or evidence, of $G$ after observing data $\mathbf{D}$ is then given by
\begin{equation}
\label{eq:marginal}
P(\mathbf{D}|G)=\int_{\Theta_G} P(\mathbf{D}|\theta_G,G)P(\theta_G|G)d\theta_G,
\end{equation}
and the posterior distributions over graphs $G$ and parameters $\theta_G$ are respectively given by 
\begin{equation}
     P(G|\mathbf{D})\propto P(G)P(\mathbf{D}|G), \quad\quad \text{and} \quad\quad
    P(\theta_G|\mathbf{D},G)\propto P(\theta_G|G)P(\mathbf{D}|\theta_G,G)\label{eq:posteriors}.
\end{equation}

This Bayesian approach lends itself naturally to our sequential setting, since the posteriors in equation \eqref{eq:posteriors} at time $t$ act as new priors for the next observation at time $(t+1)$.
One of the main challenges, however, is that the integral in equation \eqref{eq:marginal} is generally not available in closed form for arbitrary combinations of likelihood and prior.
This fact will inform our model choice later on.

While score- or constraint-based causal discovery from observational data  can usually not go beyond the Markov equivalence class, it is an advantage of the active learning approach that targeted experiments, in principle, allow to uniquely recover the causal graph.\footnote{Given infinite data, no hidden confounders, and allowing interventions on multiple variables simultaneously.}
This is because each causal graph $G\in\mathcal{G}$ implies different interventional distributions through its truncated factorisation \cite{Pearl2009},
\begin{equation}
\label{eq:ABCD-truncated-fac}
    P(\mathbf{X}_{-j}|G,do(X_j=x))=\prod_{i\neq j}P(X_i|\mathbf{Pa}_i^G)\Big\rvert_{X_j=x}.
\end{equation}
For the bivariate case of distinguishing $X\rightarrow Y$ and $Y\rightarrow X$ this is illustrated in Table \ref{tab:ABCD-interventions}.

\begin{table}[]
    \centering
    \ra{1.2}
    \caption[Different interventional distributions implied by Markov equivalent graphs]{Example of how interventional data helps to distinguish graphs within a Markov equivalence class for a setting with two variables $X,Y$. Since the two Markov equivalent graphs have different parent sets, $\mathbf{Pa}_i^{G_1}\neq\mathbf{Pa}_i^{G_2}$, they imply different interventional distributions. Experimental data is therefore used to update a marginal distribution in one, and a conditional distribution in the other.}
    \begin{tabularx}{0.75\textwidth}{@{}XXX@{}}
        \toprule
        Intervention & $G_1: X\rightarrow Y$ & $G_2: Y\rightarrow X$\\
        \midrule
        $p(y|do(x))$ & $p(y|x)$ & $p(y)$\\
        $p(x|do(y))$ & $p(x)$ & $p(x|y)$\\
       \bottomrule
    \end{tabularx}
    \label{tab:ABCD-interventions}
\end{table}

Next, we discuss how to use uncertainties over $(G,\theta_G)$ as captured by their posteriors to decide which experiment to perform next.
To tackle this optimal design problem, we turn to \textit{Bayesian experimental design} \citep{lindley1956measure, chaloner1995bayesian}.
Bayesian experimental design is a decision theoretic approach for selecting an experiment $\xi$ aiming to maximise a given utility function $U(y|\xi)$ which describes the usefulness of outcome $y$.
Given the current model specified by a prior $P(\theta)$ and a likelihood $P(y|\theta,\xi)$, the optimal experiment $\xi^*$ is the one which maximises expected utility,
\begin{equation}
\label{eq:maximum-utility}
    \xi^* = \argmax_{\xi}\int_{\mathcal{Y}} U(y|\xi) P(y|\xi) dy,
\end{equation}
where $P(y|\xi)$ is the predictive posterior for outcome $y$ under experiment $\xi$.
When the goal is to learn the parameters $\theta$, a principled utility function rooted in information theory \citep{shannon1948mathematical} is the information gain in $\theta$ from performing $\xi$ and observing $y$,
\begin{equation}
\label{eq:infogain}
    U(y|\xi)=\int_\Theta P(\theta|y,\xi)\log P(\theta|y,\xi)d\theta - \int_\Theta P(\theta)\log P(\theta)d\theta.
\end{equation}
Equivalently, equation \eqref{eq:infogain} can also be interpreted as the mutual information between $\theta$ and the experiment and its outcome $(\xi,y)$, or as the expected reduction in entropy in $\theta$ from performing $\xi$ and observing $y$. 
We will drop the second term in the following as it does not depend on $y$ or $\xi$ and is therefore irrelevant for the arg max in equation \eqref{eq:maximum-utility}.

\section{Proposed approach}
In our case, the experiment $\xi$ corresponds to an intervention of the form $do(X_j=x)$, the outcome $y$ corresponds to an observation of the remaining variables $\mathbf{X}_{-j}$, and the parameters $\theta$ correspond to the pair $(G,\theta_G)$ as both the causal graph and its functional relations are unknown.
Since we are interested in learning the the causal structure, we choose to use information gain in $G$ as utility:
\begin{equation}
\label{eq:utility-ours}
    U(\mathbf{X}_{-j}|do(X_j=x))=\sum_{G\in\mathcal{G}} P(G|\mathbf{X}_{-j},do(X_j=x)) \log P(G|\mathbf{X}_{-j},do(X_j=x)).
\end{equation}
The optimal experiment \eqref{eq:maximum-utility} for our setting can then be written as (see Appendix \ref{sec:appendix-derivation} for a derivation):
\begin{equation}
\label{eq:objective}
    (j^*,x^*) = \argmax_{j\in\{1,...,d\}, x\in \mathcal{X}_j} \sum_{G\in\mathcal{G}}P(G) \int P(\mathbf{x}_{-j}|G,do(X_j=x)) \log P(G|\mathbf{x}_{-j},do(X_j=x)) d\mathbf{x}_{-j}
\end{equation}
This reveals two major challenges for choosing good interventions:
first, we need to be able to compute (or efficiently approximate) the integral in equation \eqref{eq:objective}, which involves the predictive posterior for outcomes in $G$ and the  graph posterior after observing evidence;
and second, we need to find the arg max over $(j,x)$ where $x$ is chosen from an uncountable set.\footnote{Another challenge is the summation over all possible graphs, since the number of DAGs grows super-exponentially \cite{robinson1973counting}. Here, we think of a setting with few variables and focus on the other challenges instead.}

\begin{figure}[]
    \centering
    \includegraphics[width=\textwidth, trim={7em 15.9em 7em 1em}, clip]{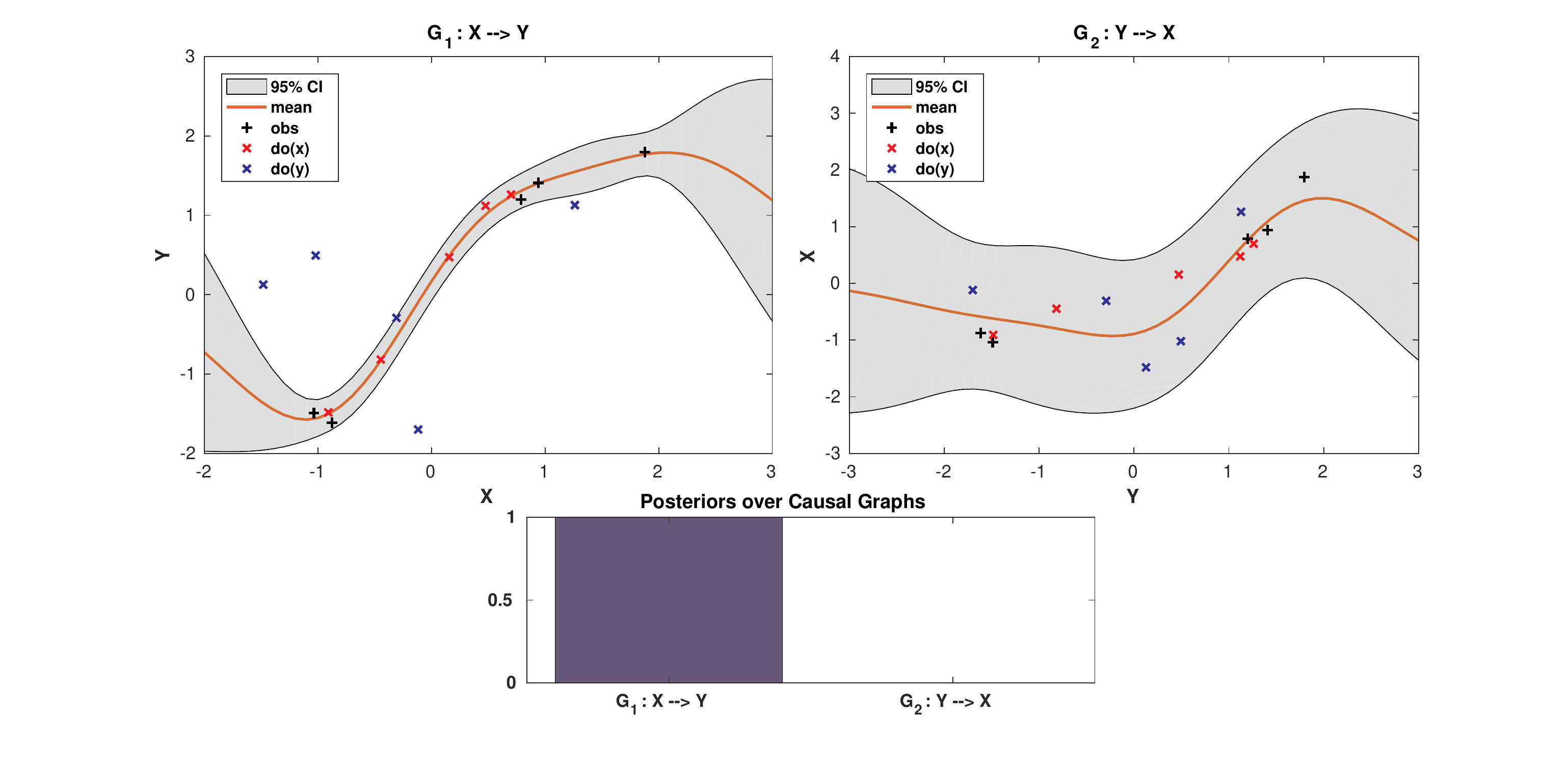}
    \caption[Prototype for Bayesian causal discovery with GP networks]{An example of a first prototype for the proposed active Bayesian causal discovery approach for a bivariate setting. Shown are the GP posteriors of the conditional distribution for $X\rightarrow Y$ (left) and $Y\rightarrow X$ (right). Plots are based on 15 samples (five observational and five from interventions on each of $X$ and $Y$) from the true data generating mechanism $X\sim \mathcal{N}(0,1)$ and $Y:=2\tanh(X)+\epsilon$, with $\epsilon\sim\mathcal{N}(0,0.1)$. Fits were initialised from observational samples, and interventions were performed alternating between $X$ and $Y$ as a target, and with intervention values drawn at random, i.e., without the proposed Bayesian optimisation scheme. After only ten such interventions, the posterior over graphs (not shown) has converged to the correct model (left) with >99\% confidence.}
    \label{fig:ABCD-prototype}
\end{figure}

We propose to tackle the first of these computational challenges while allowing for flexible non-linear relationships by using \textit{Gaussian processes} (GPs) \cite{williams2006gaussian} as priors over the functions $f_i$ in \eqref{eq:ANM}, see Figure \ref{fig:ABCD-prototype} for an illustration.
GPs are stochastic processes which can be understood as an infinite-dimensional extension of the multivariate Gaussian distribution.
They are a popular approach for nonparametric regression due to their nice analytical properties:
assuming Gaussian observation noise in \eqref{eq:ANM}, $\epsilon_i\sim\mathcal{N}(0,\sigma_i^2)$, the marginal likelihood \eqref{eq:marginal} and posteriors \eqref{eq:posteriors} are available in closed form (see Appendix \ref{sec:appendix-GPs} for details).
Because of this, we can approximate the objective \eqref{eq:objective} using a Monte Carlo estimator,
\begin{equation}
\label{eq:ABCD-Monte-Carlo-appox}
    (j^*,x^*)\approx\argmax_{j\in\{1,...,d\}, x\in \mathcal{X}_j} \sum_{G\in\mathcal{G}}P(G) \frac{1}{M} \sum_{m=1}^M \log P(G|\mathbf{x}_{-j}^{(m)},do(X_j=x)),
\end{equation}
by drawing $M$ samples $\mathbf{x}_{-j}^{(m)}$ from the interventional distribution $P(\mathbf{X}_{-j}|G,do(X_j=x))$ implied by $G$ (see Eq.~\ref{eq:ABCD-truncated-fac}) for each graph $G$.
Given a sample $\mathbf{x}_{-j}^{(m)}$, the log posterior over graphs in \eqref{eq:ABCD-Monte-Carlo-appox} can then be computed according to \eqref{eq:posteriors} using the prior over graphs and the GP marginal likelihood (see Eq.~\ref{eq:GP_marginal_likelihood} in Appendix \ref{sec:appendix-GPs}), which decomposes over the graph analogously to \eqref{eq:ABCD-truncated-fac}. 
Since we can efficiently sample from a Gaussian distribution, and since all necessary ingredients can be computed in closed form when using GPs, this overcomes the first computational challenge.\footnote{Provided that we can perform the summation over graphs, i.e., in a regime with small $d$.}

To address the second challenge, we require a smart and principled approach to compute the arg max over possible interventions in \eqref{eq:ABCD-Monte-Carlo-appox}.
For this we propose to use \textit{Bayesian optimisation} (BO) \cite{mockus1975bayesian,mockus2012bayesian}.
BO is a derivative-free technique for global optimisation which aims to solve 
\begin{equation}
\label{eq:Bayesian_optimisation}
    \argmax_{\mathbf{x}\in\mathcal{X}}f(\mathbf{x})
\end{equation}
where $f$ is typically a black-box function which is very costly to evaluate.
% In machine learning, for example, BO has become a popular approach for hyperparameter tuning in large neural networks \citep{snoek2012practical}.
The general idea is to trade off computation due to evaluating $f$ many times with computation invested in selecting more promising candidate solutions $\mathbf{x}$.
One standard approach to BO is to model uncertainty in $f$ with a GP, and to use an acquisition function $a(\mathbf{x})$ to select new candidate solutions.
A common choice for $a(\mathbf{x})$ is the upper confidence bound of the GP posterior,
\begin{equation}
\label{eq:GP-UCB}
    \mathbf{x}_{t+1}=\argmax_{\mathbf{x}} a_t(\mathbf{x}), \quad a_t(\mathbf{x})=\mu_t(\mathbf{x})+\beta \sigma_t(\mathbf{x})
\end{equation}
where $\mu_t(\mathbf{x})$ and $\sigma_t(\mathbf{x})$ correspond to the mean and standard deviation of the GP predictive distribution (see Eq.~\ref{eq:GP_predictive} in Appendix \ref{sec:appendix-GPs}) after evaluating $f$ at $\mathbf{x}_1,...,\mathbf{x}_t$, and can be interpreted as exploitation and exploration terms, respectively.
This approach is known as the GP-UCB algorithm \citep{srinivas2009gaussian}.

In our setting, we propose to run the GP-UCB algorithm $d$ times (for each possible intervention target from $\{X_1, ..., X_d\}$) to deal with the discrete optimisation variable $j$, using the Monte Carlo--approximated expected information gain in \eqref{eq:objective} as objective function.

\section{Related work}
The earliest work we are aware of on causal discovery from both observational and interventional data in a Bayesian framework is that of \citet{cooper1999causal} using perfect interventions. 
\citet{tian2001causal}, on the other hand, consider data obtained through a series of mechanism changes which affect the conditional distributions but not the parent sets.
As opposed to the previous two works, \citet{eaton2007exact} assume that intervention targets are not known a priori, and propose to learn both targets and causal graph using the dynamic programming approach of \citet{koivisto2004exact}.

Most closely related to our proposed approach are the works of \citet{murphy2001active} and \citet{tong2001active} who, rather than learning from passively obtained interventional data, consider the task of actively choosing interventions using an information theoretic approach.
All of these exclusively consider discrete variables though, assuming a conjugate Dirichlet-multinomial model for computational convenience. 
Only recently has this approach been extended to continuous variables \citep{cho2016reconstructing,ness2017bayesian,agrawal2019abcd}, though limited to linear Gaussian models, and considering only a finite number (rather than a continuous range) of possible intervention values.

A different line of work approaches the experimental design problem from a theoretical perspective on the graph-level \citep{eberhardt2008almost, eberhardt2006n,hauser2014two, he2008active,ghassami2017budgeted}.
While providing useful insights, these intervention strategies are usually designed for the distribution level (i.e., assuming infinite data), and remain very general in that they do not assume a specific model.
Instead, we consider the finite-data setting where each experiment only provides one (or a few) samples. 

To the best of our knowledge, ours is the first work to address the Bayesian experimental design problem for continuous variables, allowing for flexible non-linear relationships through the use of GP networks \citep{friedman2000gaussian}.
Unlike other active learning schemes for GP networks \cite{rubenstein2017probabilistic}, we aim to learn both functional relationships and  network structure simultaneously.
This requires taking into account both types of uncertainty (i.e., over functions and graphs) when selecting new interventions.

\section{Discussion and open problems}
\label{ch:5-ABCD-discussion}
Finally, we discuss some issues and open problems of the proposed approach which we have not covered so far, and suggest ideas and possible solutions for addressing these in future work.

\paragraph{Dirichlet process mixtures of Gaussians as flexible input distributions}
In absence of parents as inputs, GP networks in their original form \citep{friedman2000gaussian} use simple Gaussians as marginal distributions for root nodes.
However, this is not in the spirit of allowing for flexible relationships as the complexity of a system depends on both the input and the mechanism \citep{janzing2012information}:
a complex joint distribution may arise from either a simple marginal and a complex mechanism, or from a complex marginal and a simple mechanism.
To allow for the latter to be captured by our model, we thus suggest to model the marginals of root nodes using Dirichlet mixtures of Gaussians as a flexible Bayesian non-parametric density estimator \citep{ferguson1973bayesian}.
While this poses a computational challenge, we suggest to use an efficient variational inference technique for Dirichlet process mixture models proposed by \citet{blei2006variational}.

\paragraph{Adaptive updating of hyperparameters}
Another issue is how to choose the GP hyperparameters (i.e., lengthscales, signal variances, and noise variances) for each variable and each graph.
The standard approach for this is to take a type-2 maximum likelihood approach which selects hyperparameters by maximising the marginal likelihood \citep{williams2006gaussian}.
While this approach works very well in an offline learning setting, it may be less suitable for our online setting.
This is because updating hyperparameters at each step is both a computational burden and changes the model which makes iterative model comparison difficult.
An alternative is to take a fully Bayesian approach by placing a hyperprior over hyperparameters and to update their posterior at each step.
This has the disadvantage that the marginal likelihood is no longer available in closed form, and performing approximate inference thus becomes necessary.
When a sufficiently large observational sample is available initially, these problems can be avoided by using it to fit the hyperparameters and keeping them fixed thereafter.

\paragraph{On the choice of graph prior}
An advantage of the Bayesian approach is that available domain knowledge may be incorporated into the prior.
For example, when using our framework for experimental design in a scientific setting, we may have access to a reference graph which captures current expert beliefs.
In this case, setting the prior probability of each graph to be inversely proportional to its structural distance to the reference graph seems to be a good choice.
Alternatively, the Markov equivalence class may be known from observational data, drastically reducing the number of graphs which need to be considered. 
Yet another approach is to enforce sparsity via the prior, e.g., by setting the prior probability to be inversely proportional to the number of edges, or even fixing a maximum number of edges per graph or node and setting the prior equal to zero for any graph exceeding this number. 
All the above may help to reduce the computational burden by restricting the set of DAGs under consideration.
If no background knowledge is available, a uniform prior can be used.

\paragraph{On the combinatorial number of DAGs}
An important aspect of the proposed framework is the computational burden introduced by the super-exponential number of DAGs \citep{robinson1973counting}.
While this can be addressed through sampling approaches such as MCMC in the space of graphs or topological orderings \citep{friedman2003being}, we consider this issue  as orthogonal to the challenges we aim to address here, i.e., dealing with continuous variables with non-linear relationships and uncountable intervention sets.
To study these problems in isolation we suggest to focus on settings with only a few observed variables (i.e., small $d$) where it is still possible to simply enumerate all DAGs. 
To scale our approach up to high-dimensional problems, however, it will eventually be necessary to combine it with efficient approximate inference procedures, such as the minimal I-map MCMC of \citet{agrawal2018minimal}.

\paragraph{Relation to and combination with other causal discovery techniques}
The nonlinear ANM we assume in \eqref{eq:ANM} also, in principle, allows to identify causal structure from purely observational data as shown by \citet{hoyer2009nonlinear}.
However, their approach relies on testing the regression residuals for independence, and therefore requires a sizeable amount of data to achieve statistical significance.
Our Bayesian score-based approach, on the other hand, is aimed at drawing maximal insights from very limited amounts of data by actively choosing where to intervene next.
In such a small data-regime, the well-known poor scaling of GPs (cubic in the number of observations) is less problematic---the computational bottleneck in our setting is instead given by the inner loop of Monte Carlo sampling combined with BO (as well as the number of DAGs as mentioned before).
To improve tractability, a combination of our method with constraint-based approaches \cite{Spirtes1993}, and regression with subsequent independence-testing (RESIT) \citep{peters2014causal} seems to be an interesting direction for future work.

\paragraph{Applicability and suggested use cases}
Due to the aforementioned computational challenges, our proposed approach in its current form may be particularly useful for situations (i) involving only small number of observed variables, and (ii) in which experiments are expensive compared to computation time. 
This suggests decision support for experimental scientists as one suitable use case.
There, the involved variables are often continuous and exhibit non-linear (e.g., saturating) behaviour, experiments are usually very expensive to perform and results may take a long time to arrive, and the heavy use of computational resources to increase scientific insights gained from wisely targeted experiments is therefore often justified.

\clearpage
\bibliographystyle{unsrtnat}
\bibliography{references.bib}

\appendix
\section{Derivation of expected information gain}
\label{sec:appendix-derivation}
% The predictive posterior for our setting takes the form
% \begin{equation}
% \label{eq:predictive-posterior-ours}
%     P(\mathbf{X}_{-j}|do(X_j=x))=\sum_{G\in\mathcal{G}}P(G)P(\mathbf{X}_{-j}|G,  do(X_j=x)),
% \end{equation}
% which can be seen as a weighted average of predictions according to our current beliefs.\footnote{Recall, that the priors for later time points correspond to the posteriors at the previous time point.}

The optimal intervention according to \eqref{eq:maximum-utility} is given by
\begin{equation}
\label{eq:optimal-intervention}
    (j^*,x^*) = \argmax_{j\in\{1,...,d\}, x\in \mathcal{X}_j} \int U(\mathbf{x}_{-j}|do(X_j=x)) P(\mathbf{x}_{-j}|do(X_j=x))d\mathbf{x}_{-j}
\end{equation}
By substituting \eqref{eq:utility-ours} into \eqref{eq:optimal-intervention} we can rewrite this integral as
\begin{align}
    & \int  \sum_{G\in\mathcal{G}} P(G|\mathbf{x}_{-j},do(X_j=x)) \log P(G|\mathbf{x}_{-j},do(X_j=x)) P(\mathbf{x}_{-j}|do(X_j=x))d\mathbf{x}_{-j}\\
      = & \sum_{G\in\mathcal{G}}\int P(G,\mathbf{x}_{-j}|do(X_j=x)) \log P(G|\mathbf{x}_{-j},do(X_j=x)) d\mathbf{x}_{-j}\\
    = & \sum_{G\in\mathcal{G}}P(G) \int P(\mathbf{x}_{-j}|G,do(X_j=x)) \log P(G|\mathbf{x}_{-j},do(X_j=x)) d\mathbf{x}_{-j} \label{eq:ABCD-integral-final-form}
\end{align}
where the last line coincides with \eqref{eq:objective}.
The reason for writing the objective in this form is that the GP predictive posterior and marginal likelihood have a closed form when conditioning on the graph $G$, but not otherwise.

\section{Background on Gaussian processes}
\label{sec:appendix-GPs}
\begin{definition}[GP]
A Gaussian process (GP) is a collection of random variables, any finite number of which have a joint Gaussian distribution.
\end{definition}

Just like the Gaussian distribution, a GP is fully determined by its mean and covariance. We write $f\sim GP(m,k)$ to denote that the process $f(\mathbf{x})$ is a GP with mean function $m$ and covariance function (or kernel) $k$, where 
\begin{equation}
m(\mathbf{x})=\mathbb{E}[f(\mathbf{x})], \quad \text{and} \quad  k(\mathbf{x},\mathbf{x}')=\mathbb{E}[(f(\mathbf{x})-m(\mathbf{x}))(f(\mathbf{x}')-m(\mathbf{x}'))].
\end{equation}
The mean function is usually taken to be zero, $m(x)\equiv 0$, while a common choice of covariance function is given by the squared-exponential,
\begin{equation}\label{eq:squared_exponential}
k_{SE}(\mathbf{x},\mathbf{x}')=\lambda\exp(-\sum_{i=1}^p\nu_i(x_i-x_i')^2).
\end{equation}
Here, the signal variance $\lambda$ and inverse length scales $\nu_i$ are considered hyperparameters.

What makes inference with GPs possible is that we only ever reason about function values at a finite set of locations. Consider a regression model with observation noise,
\begin{equation}
   y = f(\mathbf{x}) + \epsilon, \quad \text{where} \quad \epsilon \sim~\mathcal{N}(0,\sigma^2) 
\end{equation}
so $y|f, \mathbf{x}\sim \mathcal{N}(f(\mathbf{x}),\sigma^2)$, and assume a GP prior $f\sim GP(0,k)$.
Given $n$ observations $\mathbf{X}=(\mathbf{x}_1, ..., \mathbf{x}_n)$ and $\mathbf{y}=(y_1,...,y_n)$ denote by $\mathbf{f}=(f(\mathbf{x}_1),..., f(\mathbf{x}_n))$ the vector of function values and by $K$ the matrix with entries $K_{ij}=k(\mathbf{x}_i, \mathbf{x}_j)$, also called the Gram matrix.
By definition, $\mathbf{f}|\mathbf{X}\sim\mathcal{N}(0,K)$, and using Gaussian identities \citep[see e.g.,][Appendix A.2 for details]{williams2006gaussian} we can compute the log marginal likelihood in closed form:
\begin{equation}
\label{eq:GP_marginal_likelihood}
    \log p(\mathbf{y}|\mathbf{X}) = \log \int p(\mathbf{y}|\mathbf{f},\mathbf{X})p(\mathbf{f}|\mathbf{X})d\mathbf{f}
    = -\frac{1}{2}\mathbf{y}^T(K+\sigma I)^{-1}\mathbf{y}-\frac{1}{2}\log |K+\sigma I| -\frac{n}{2}\log2\pi 
\end{equation}
To predict the function value $f_*$ at a new location $\mathbf{x}_*$, we observe that $(\mathbf{y}, f_*)$ has a joint Gaussian distribution.
Denote by $\mathbf{k}_*=(k(\mathbf{x}_1, \mathbf{x}_*),...,k(\mathbf{x}_n, \mathbf{x}_*))$ the vector of covariances between $\mathbf{x}_*$ and the $n$ observations. By conditioning on $\mathbf{y}$ we obtain the predictive posterior,
\begin{equation}
\label{eq:GP_predictive}
    f_*|\mathbf{x}_*, \mathbf{y}, \mathbf{X} \sim
    \mathcal{N}(\mathbf{k}_*^T(K+\sigma^2I)^{-1}\mathbf{y}, k(\mathbf{x}_*,\mathbf{x}_*)-\mathbf{k}_*^T(K+\sigma^2I)^{-1}\mathbf{k}_*).
\end{equation}

Equations \eqref{eq:GP_marginal_likelihood} and \eqref{eq:GP_predictive} provide closed form expressions for the marginal likelihood (needed to compute the posterior over graphs) and the predictive posterior (needed to predict experimental outcomes).
This computational tractability makes the use of GPs attractive for our setting.
Concretely, we consider the following model within each graph $G\in\mathcal{G}$:
\begin{equation}
\label{eq:ABCD-model}
    X_i = f_i(\mathbf{Pa}_i^G)+\epsilon_i, \quad f_i\sim GP(0,k_{SE}), \quad \epsilon_i \sim \mathcal{N}(0,\sigma_i), \quad \text{for} \quad i=1,...,d.
\end{equation}

\end{document}